\documentclass[10pt,twocolumn,letterpaper]{article}

\usepackage{cvpr}
\usepackage{times}
\usepackage{epsfig}
\usepackage{graphicx}
\usepackage{amsmath}
\usepackage{amssymb}
\usepackage{adjustbox}
\usepackage{booktabs}
\usepackage{siunitx}

\usepackage[pagebackref=true,breaklinks=true,bookmarks=false, colorlinks]{hyperref}

\makeatletter
\def\@fnsymbol#1{\ensuremath{\ifcase#1\or \dagger \or *\or \ddagger\or
\mathsection\or \mathparagraph\or \|\or **\or \dagger\dagger
\or \ddagger\ddagger \else\@ctrerr\fi}}
\makeatother

\cvprfinalcopy 

\def\cvprPaperID{1785} 

\ifcvprfinal\fi
\begin{document}

\title{Self-supervised Spatio-temporal Representation Learning for Videos \\by Predicting Motion and Appearance Statistics}

\author{Jiangliu Wang$^1 \thanks{Work done during an internship at Tencent AI Lab.}$ ~~~ Jianbo Jiao$^{2\dagger}$ ~~~  Linchao Bao$^3 \thanks{Corresponding authors.}$ ~~~ Shengfeng He$^4$ ~~~ Yunhui Liu$^1$ ~~~ Wei Liu$^{3*}$ \\
$^1$The Chinese University of Hong Kong \quad $^2$University of Oxford \\
$^3$Tencent AI Lab\quad $^4$South China University of Technology }

\maketitle
\thispagestyle{empty}

\begin{abstract}
We address the problem of video representation learning without human-annotated labels. 
While previous efforts address the problem by designing novel self-supervised tasks using video data, the learned features are merely on a frame-by-frame basis, which are not applicable to many video analytic tasks where spatio-temporal features are prevailing. 
In this paper we propose a novel self-supervised approach to learn spatio-temporal features for video representation. 
Inspired by the success of two-stream approaches in video classification, we propose to learn visual features by regressing both motion and appearance statistics along spatial and temporal dimensions, given only the input video data. 
Specifically, we extract statistical concepts (fast-motion region and the corresponding dominant direction, spatio-temporal color diversity, dominant color, \etc) from simple patterns in both spatial and temporal domains. Unlike prior puzzles that are even hard for humans to solve, the proposed approach is consistent with human inherent visual habits and therefore easy to answer.
We conduct extensive experiments with C3D to validate the effectiveness of our proposed approach. 
The experiments show that our approach can significantly improve the performance of C3D when applied to video classification tasks. 
Code is available at \href{https://github.com/laura-wang/video_repres_mas}{https://github.com/laura-wang/video\_repres\_mas}.
\end{abstract}

\section{Introduction}

Learning powerful spatio-temporal representations is the most fundamental deep learning problem for many video understanding tasks such as action recognition \cite{carreira2017quo, jiang2015human, liu2016video}, action proposal and localization \cite{chao2018rethinking, shou2017cdc, shou2016temporal}, video captioning \cite{wang2018reconstruction, wang2018bidirectional}, \etc ~
Great progresses have been made by training expressive networks with massive human-annotated video data \cite{tran2015learning,r2plus1d_cvpr18}. 
However, annotating video data is very laborious and expensive, which makes the learning from unlabeled video data important and interesting.

\begin{figure}[t]
    \centering
    \includegraphics[width=\columnwidth]{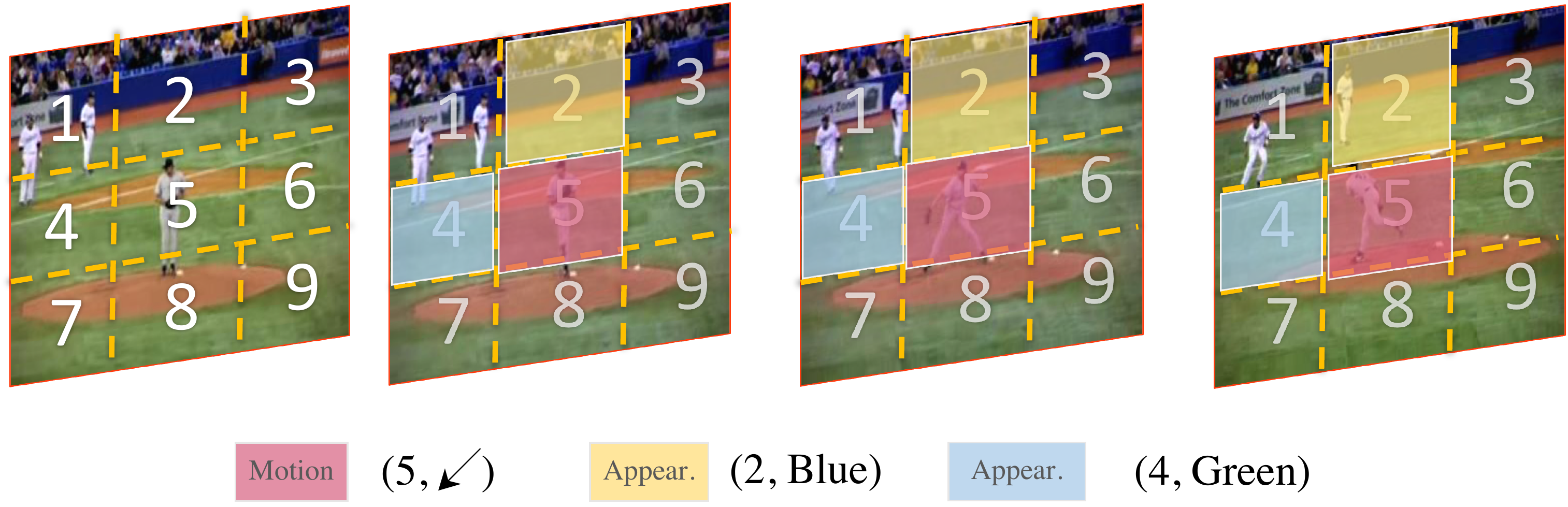}
    \caption{The main idea of the proposed approach. Given a video sequence, we design a novel task to predict several numerical labels derived from motion and appearance statistics for  spatio-temporal representation learning, in a self-supervised manner. Each video frame is first divided into several spatial regions using different partitioning patterns like the grid shown above. Then the derived statistical labels, such as \emph{the region with the largest motion and its direction} (the red patch), \emph{the most diverged region in appearance and its dominant color} (the yellow patch), and \emph{the most stable region in appearance and its dominant color} (the blue patch), are employed as supervision during the learning.}
    \label{fig:teas}\vspace{-4mm}
\end{figure}

Recently, several approaches \cite{misra2016shuffle, fernando2017self, lee2017unsupervised, gan2018geometry} have emerged to learn transferable representations for video recognition tasks with unlabeled video data. 
In these approaches, a CNN is first pre-trained on unlabeled video data using novel self-supervised tasks, where supervision signals can be easily derived from input data without human labors, such as solving puzzles with perturbed video frame orders \cite{misra2016shuffle,fernando2017self,lee2017unsupervised} or predicting flow fields or disparity maps obtained with other computational approaches \cite{gan2018geometry}. 
Then the learned representations can be directly applied to other video tasks as features, or be employed as initialization during succeeding supervised learning. 
Unfortunately, although these work demonstrated the effectiveness of self-supervised representation learning with unlabeled videos, their approaches are only applicable to a CNN that accepts one or two frames as inputs, which is not a recommended way for tackling video tasks. 
In most video understanding tasks, spatio-temporal features that can capture information of both appearances and motions are proved to be vital in many recent studies \cite{cao2013mining,simonyan2014two,tran2015learning,carreira2017quo,r2plus1d_cvpr18}. 

In order to extract spatio-temporal features, a network architecture that can accept multiple frames as inputs and perform operations along both spatial and temporal dimensions is needed. 
For example, the popular C3D network \cite{tran2015learning}, which accepts 16 frames as inputs and employs 3D convolutions along both spatial and temporal dimensions to extract features, is becoming more and more popular for many video tasks \cite{shou2017cdc, shou2016temporal, krishna2017dense, li2018jointly, wang2018bidirectional}. Vondrick \etal \cite{vondrick2016generating} proposed to address the representation learning by C3D-based networks, while motion and appearance are not explicitly incorporated thus the performance is not satisfactory when transferring the learned features to other video tasks. 

In this paper, we propose a novel self-supervised learning approach to learn spatio-temporal video representations by predicting motion and appearance statistics in unlabeled videos.
The idea is inspired by Giese and Poggio's work on human visual system \cite{giese2003cognitive}, in which the representation of motion is found to be based on a set of learned patterns. 
These patterns are encoded as sequences of ‘snapshots’ of body shapes by neurons in the \emph{form pathway}, and by sequences of complex optic flow patterns in the \emph{motion pathway}. 
In our work, the two pathways are the appearance branch and motion branch respectively. Besides, the abstract statistical concepts are also inspired by the biological hierarchical perception mechanism.
The main idea of our approach is shown in Figure~\ref{fig:teas}.
We design several spatial partitioning patterns to encode each spatial location and its motion and appearance statistics over multiple frames, and use the encoded vectors as supervision signals to train the spatio-temporal representation network. 
The novel objectives are simple to learn and informative for the motion and appearance distributions in video, \eg, the spatial locations of the most dominant motions and their directions, the most consistent and the most diverse colors over a certain temporal cube, 
\etc ~ 
We conduct extensive experiments with C3D network to validate the effectiveness of the proposed approach. We show that, compared with training from scratch, pre-training C3D without labels using our proposed approach gives a large boost to the performance of the action recognition task (\eg, $45.4\%$ \emph{v.s.} $61.2\%$ on UCF101).
By transferring the learned representations to other video tasks on smaller datasets, we demonstrate significant performance gains on various tasks like dynamic scene recognition, action similarity labeling, \etc

\section{Related work}

Self-supervised representation learning is proposed to leverage the huge amounts of unlabeled data to learn useful representations for various problems, for example, image classification, object detection, video recognition, \etc~
It has been proved that lots of deep learning methods can benefit from pre-trained models on large labeled datasets, \eg, ImageNet \cite{deng2009imagenet} for image tasks and Kinetics \cite{kay2017kinetics} or Sports-1M \cite{karpathy2014large} for video tasks. 
The basic motivation behind self-supervised representation learning is to replace the expensive labeled data with ``free'' unlabeled data. 

A common way to achieve self-supervised learning is to derive easy-to-obtain supervision signals without human annotations, to encourage the learning of useful features for regular tasks.
Various novel tasks are proposed to learn image representations from unlabeled image data, \eg, re-ordering perturbed image patches \cite{doersch2015unsupervised,noroozi2016unsupervised}, colorizing grayscale images \cite{zhang2016colorful}, inpainting missing regions \cite{pathak2016context}, counting virtual primitives \cite{noroozi2017representation}, classifying image rotations \cite{gidaris2018unsupervised}, predicting image labels obtained using a clustering algorithm \cite{caron2018deep}, \etc. 
There are also studies that try to learn image representations from unlabeled video data. 
Wang and Gupta \cite{wang2015unsupervised} proposed to derive supervision labels from unlabeled videos using traditional tracking algorithms. 
Pathak \etal \cite{pathak2017learning} instead obtained labels from videos using conventional motion segmentation algorithms. 

Recent studies leveraging video data try to learn transferable representations for video tasks. 
Misra \etal \cite{misra2016shuffle} designed a binary classification task and asked the CNN to predict whether the video input is in right order or not. Fernando \etal \cite{fernando2017self} and Lee \etal \cite{lee2017unsupervised} also designed tasks based on video frame orders. 
Gan \etal  \cite{gan2018geometry} proposed a geometry-guided network that force the CNN to predict flow fields or disparity maps between two input frames. Although these work demonstrated the effectiveness of self-supervised representation learning with unlabeled videos and showed impressive performances when transferring the learned features to video recognition tasks, their approaches are only applicable to a CNN that accepts one or two frames as inputs and cannot be applied to network architectures that are suitable for spatio-temporal representations. 
The most related work to ours are Vondrick \etal \cite{vondrick2016generating} and Kim \etal \cite{kim2018self}. 
Vondrick \etal \cite{vondrick2016generating} proposed a GAN model for videos with a spatio-temporal 3D convolutional architecture, which can be employed as a self-supervised approach for video representation learning.
Kim \etal \cite{kim2018self} proposed to learn spatio-temporal representations with unlabeled video data, by solving space-time cubic puzzles, which is a straightforward extension of the 2D puzzles \cite{noroozi2016unsupervised}.


\section{Our Approach}

We design a novel task for self-supervised video representation learning by predicting the motion and appearance statistics in a video sequence. 
The task is bio-inspired and consistent with human visual habits \cite{giese2003cognitive} to capture high-level concepts of videos.
In this section, we first illustrate the statistical concepts and motivations to design the task (Sec. \ref{sec.concept}). 
Next, we formally define the proposed statistical labels (Sec. \ref{sec.motionstat} and \ref{sec.rgbstat}). 
Finally, we present the whole learning framework when applying the self-supervised task to the C3D \cite{tran2015learning} network (Sec. \ref{sec.learnc3d}).

\subsection{Statistical Concepts}\label{sec.concept}

Given a video clip, humans usually first notice the moving proportion of the visual field \cite{giese2003cognitive}. By observing the foreground motion and the background appearance, we can easily tell the motion class based on prior knowledge. 
Inspired by human visual system, we break the process of understanding videos into several questions and encourage a CNN to answer them accordingly: (1) Where is the largest motion in the video? (2) What is the dominant direction of the largest motion? (3) Where is the largest color diversity and what is its dominant color? (4) Where is the smallest color diversity, \ie, the potential background of a scene and what is its dominant color? 
The approach to quantify these questions into annotation-free training labels will be described in details in the following sections.
Here, we introduce the statistical concepts for motion and appearance. 
 
\begin{figure}[t]
\vspace{-5pt}
\begin{center}
   \includegraphics[width=1\linewidth]{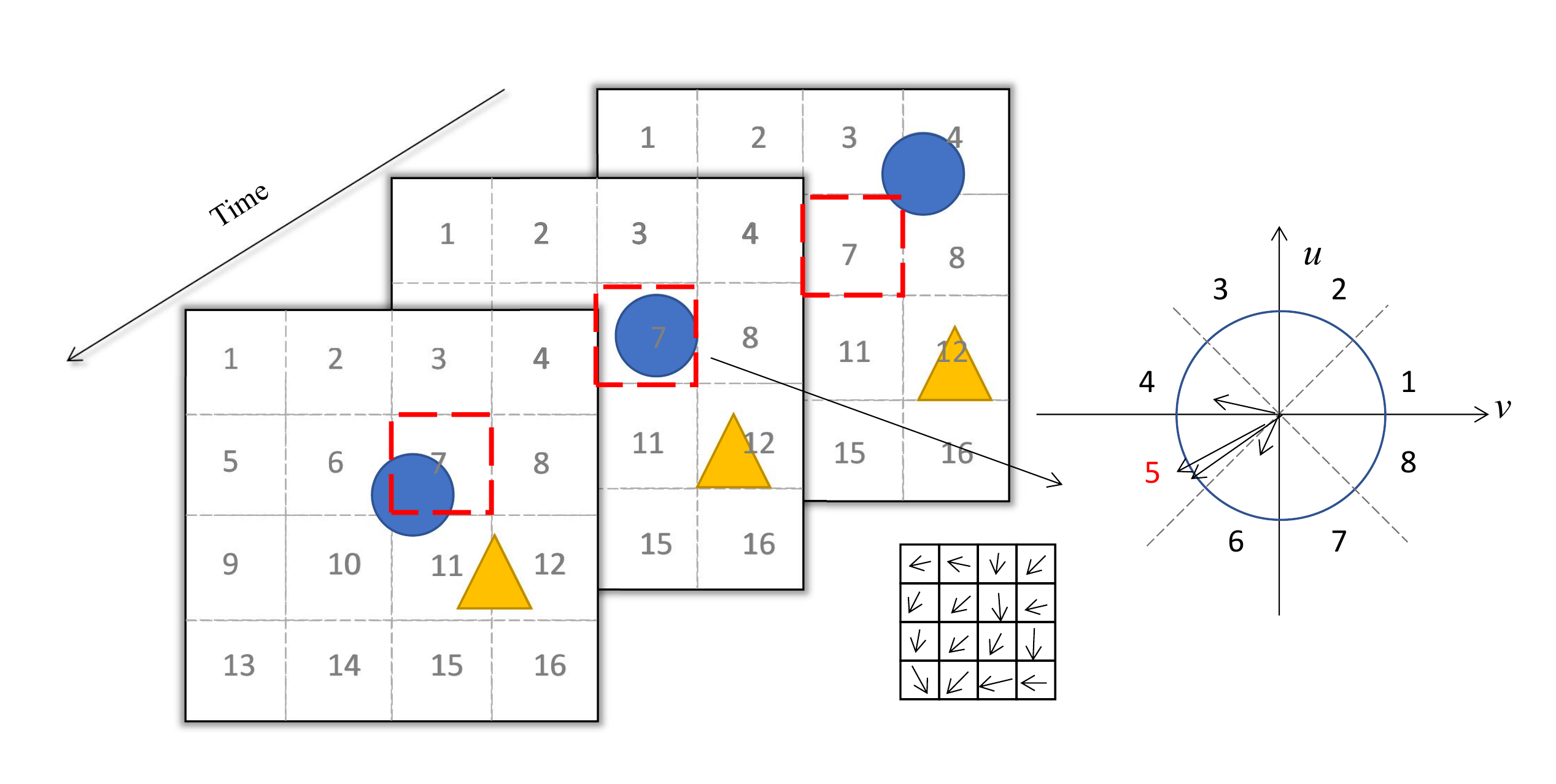}\vspace{-6mm}
\end{center}
   \caption{A simple illustration of statistical concepts in a three-frame video clip. See explanations in Sec. \ref{sec.concept} for more details.
   }
\label{fig:concepts}\vspace{-2mm}
\vspace{-5pt}
\end{figure}

Figure \ref{fig:concepts} illustrates an example of a three-frame video clip with two moving objects (blue circle and yellow triangle). 
A typical video clip normally consists of much more frames. We here instead use the three-frame clip for better understanding of the key ideas. 
To accurately represent the location and quantify ``where'', each frame is divided into 4-by-4 blocks and each block is assigned to a number in an ascending order. The blue circle moves from block four to block seven, and the yellow triangle moves from block 12 to block 11. 
Comparing the moving distance, we can easily tell that the motion of the blue circle is larger than the motion of the yellow triangle. 
And the largest motion lies in block seven since it contains moving-in motion between frame one and two, and moving-out motion between frame two and three. 
As for the question ``\emph{what is the dominant direction of the largest motion?}'', it can be easily observed from Figure \ref{fig:concepts} that the blue circle is moving towards lower-left. 
To quantify the directions, the full angle \ang{360} is divided into eight angle pieces, with each piece covering a \ang{45} motion direction range. 
And similar to location quantification, each angle piece is assigned to a number in an ascending order counterclockwise. 
The corresponding angle piece number of ``lower-left'' is five.

For the appearance statistics, the largest spatio-temporal color diversity area is also block seven, as it changes from the background color to the circle color. 
The dominant color is the same as the moving circle color, \ie, blue. 
As for the smallest color diversity location, most of the blocks stay the same and the background color is white.

Keeping the above concepts and motivation in mind, we next present the proposed novel self-supervised approach. 
We assume that by training a spatio-temporal CNN to predict the motion and appearance statistics mentioned above, better spatio-temporal representations can be learned, by which the video understanding tasks could be benefited consequentely.  
Specifically, we design a novel regression task to predict a group of numbers related to motion and appearance statistics, such that by correctly predicting them, the following queries could be roughly derived: the largest motion location and the dominant motion direction in the video, the most consistent colors over the frames and their spatial locations, and the most diverse colors over the frames and their spatial locations.

\subsection{Motion Statistics}\label{sec.motionstat}
\label{motion_s}
\vspace{-4pt}

We use optical flow computed by classic coarse-to-fine algorithms \cite{brox2004high} to derive the motion statistical labels to be predicted in our task. 
Optical flow is a motion representation feature that is commonly used in many video recognition methods. 
For example, the classic two-stream network \cite{simonyan2014two} and the recent I3D network \cite{carreira2017quo}, both of which use stack of optical flow as their inputs for action recognition tasks. 
However, optical flow based methods are sensitive to camera motion, since they represent the absolute motion \cite{carreira2017quo, wang2011action}. 
To suppress the influence of camera motion, we instead seek a more robust feature, motion boundary \cite{dalal2006human}, to capture  the video motion information.

\begin{figure*}[t]
    \centering
   \includegraphics[width=0.9\textwidth]{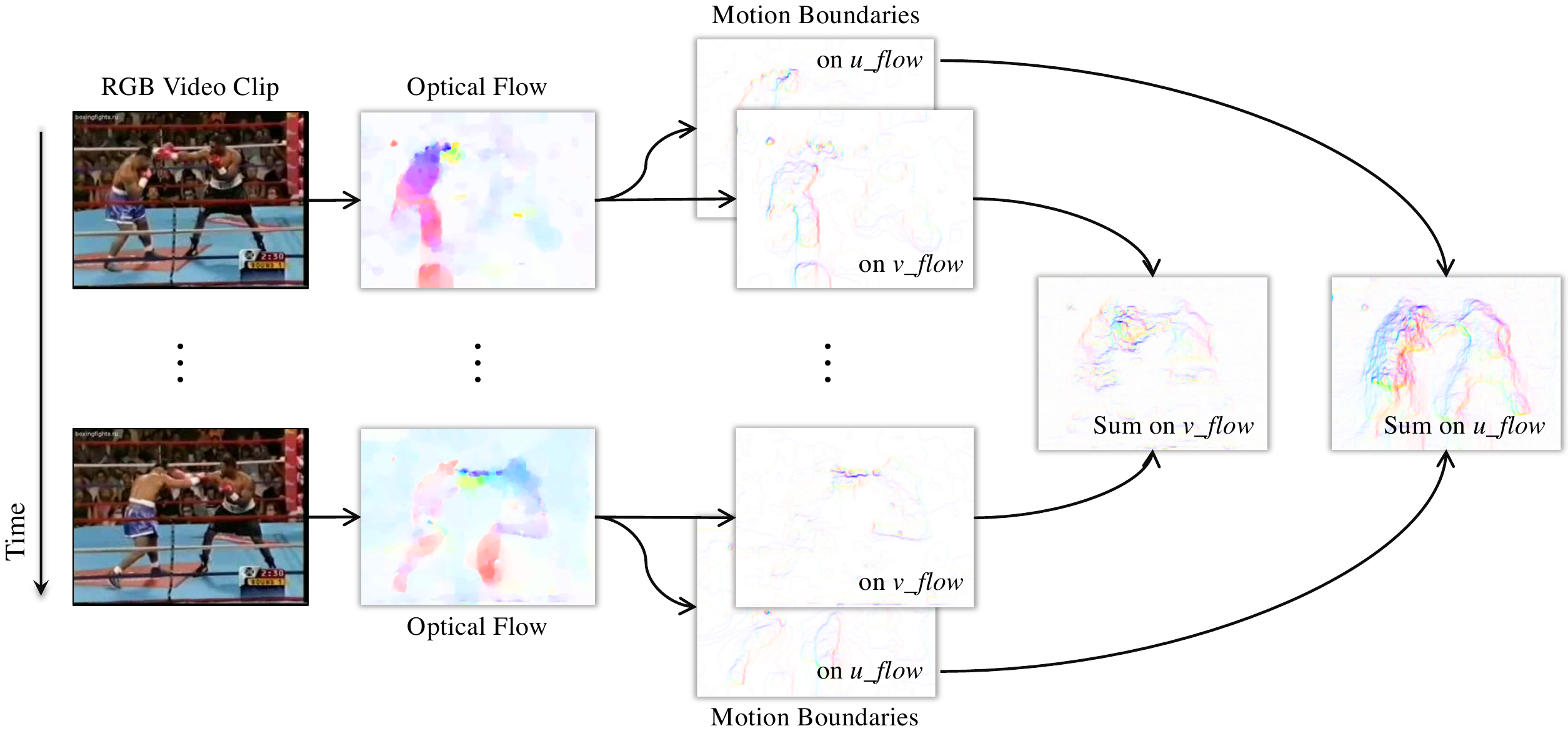}
   \caption{Motion boundaries computation. For a given input video clip, we first extract optical flow across each frame. For each optical flow, two motion boundaries are obtained by computing gradients separately on the horizontal and vertical components of the optical flow. The final sum-up motion boundaries are obtained by aggregating the motion boundaries on \emph{u\_flow} and \emph{v\_flow} of each frame separately.}\vspace{1mm}
\label{fig:motion_boudary}
\end{figure*}

\vspace{-3mm}
\paragraph{Motion Boundary.} 
Denote optical flow horizontal component and vertical component as $u$ and $v$, respectively. 
Motion boundaries are calculated by computing x- and y- derivatives of $u$ and $v$, \ie, $u_x=\frac{\partial u}{\partial x}$, $u_y=\frac{\partial u}{\partial y}$, $v_x=\frac{\partial v}{\partial x}$, $v_y=\frac{\partial v}{\partial y}$. 
As motion boundaries capture changes in the flow field, constant or smoothly varied motion, such as motion caused by camera view change, will be cancelled out. Only motion boundaries information is kept, as shown in Figure \ref{fig:motion_boudary}. 
Specifically, for an $N$-frame video clip,  $(N-1)*2$ motion boundaries are computed. 
Diverse video motion information can be encoded into two summarized motion boundaries by summing up all these $(N-1)$ sparse motion boundaries of each component as follows:
\begin{equation}
    M_u=( \sum\limits_{i=1}^{N-1}u_x^i, \sum\limits_{i=1}^{N-1}u_y^i), ~
    M_v=( \sum\limits_{i=1}^{N-1}v_x^i, \sum\limits_{i=1}^{N-1}v_y^i),
  \label{sum_up}
\end{equation}
where $M_u$ denotes the motion boundaries on horizontal optical flow $u$, and $M_v$ denotes the motion boundaries on vertical optical flow $v$. 
Figure \ref{fig:motion_boudary} shows the visualization of the two sum-up motion boundaries images.

\begin{figure}
    \centering
    \includegraphics[width=\columnwidth]{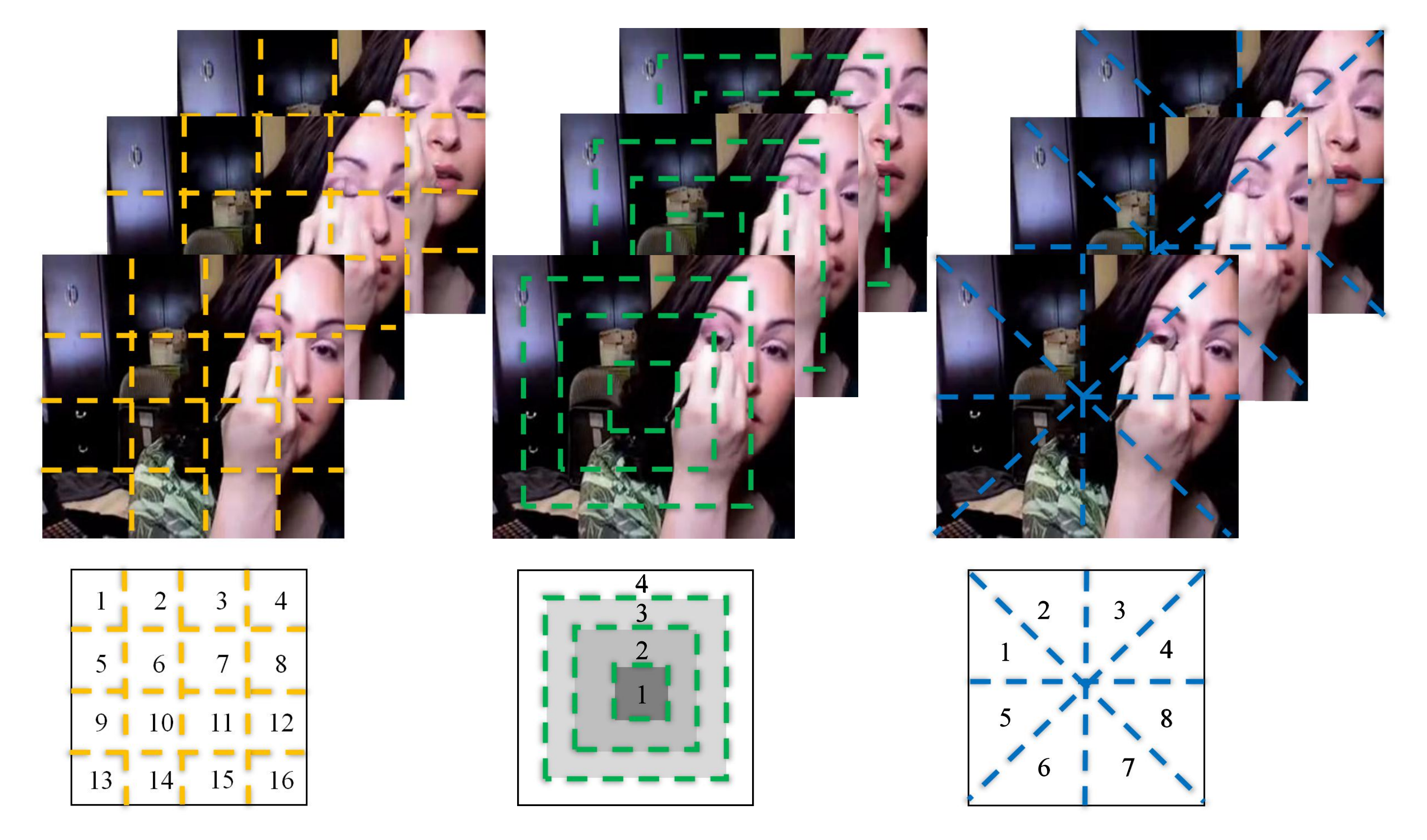}
    \caption{Three different partitioning patterns (from left to right: 1 to 3) used to divide video frames into different types of spatial regions. Pattern 1 divides each frame into 4$\times$4 blocks. Pattern 2 divides each frame into 4 different non-overlapped areas with the same gap between each block. Pattern 3 divides each frame by the two center lines and the two diagonal lines. The indexing strategies of the labels are shown in the bottom row. }
    \label{fig:pattern}\vspace{-3mm}
\end{figure}

\paragraph{Spatial-aware Motion Statistical Labels.} In this section, we describe how to design the spatial-aware motion statistical labels to be predicted by our self-supervised task: 1) where is the largest motion; 2) what is the dominant orientation of the largest motion, based on motion boundaries. 
Given a video clip, we first divide it into several blocks using simple patterns. 
Although the pattern design is an interesting problem to be investigated, here, we introduce three simple yet effective patterns as shown in Figure \ref{fig:pattern}. 
For each video block, we assign a number to it for representing its location. 
Then we compute $M_u$ and $M_v$ as described above.
The motion magnitude and orientation of each pixel can be obtained by casting motion boundaries $M_u$ and $M_v$ from the Cartesian coordinates to the Polar coordinates. 
As for the largest motion statistics, we compute the average magnitude of each block and use the number of the block with the largest average magnitude as the largest motion location.  
Note that the largest block number computed from $M_u$ and $M_v$ can be different. 
Therefore, we use two labels to represent the largest motion locations of $M_u$ and $M_v$ separately. 
While for the dominant orientation statistics, an orientation histogram is computed based on the largest motion block, similar to the computation motion boundary histogram (MBH) \cite{dalal2006human}. 
Note that we do not have the normalization step since we are not computing a descriptor. 
Instead, we divide \ang{360} into 8 bins, with each bin containing  \ang{45} angle range and again assign each bin to a number to represent its orientation. 
For each pixel in the largest motion block, we first use its orientation angle to determine which angle bin it belongs to and then add the corresponding magnitude number into the angle bin. 
The dominant orientation is the number of the angle bin with the largest magnitude sum.   

\paragraph{Global Motion Statistical Labels.} We also propose a set of global motion statistical labels to provide complementary information to the local motion statistics described above. 
Instead of focusing on the local patch of video clips, a CNN is asked to predict the largest motion frame. 
That is given an $N$-frame video clip, the CNN is encouraged to understand the video evolution from a global perspective and find out between which two frames, contains the largest motion. 
The largest motion is quantified by $M_u$ and $M_v$ separately and two labels are used to represent the global motion statistics.

\begin{figure*}
    \centering
    \includegraphics[width=1\textwidth]{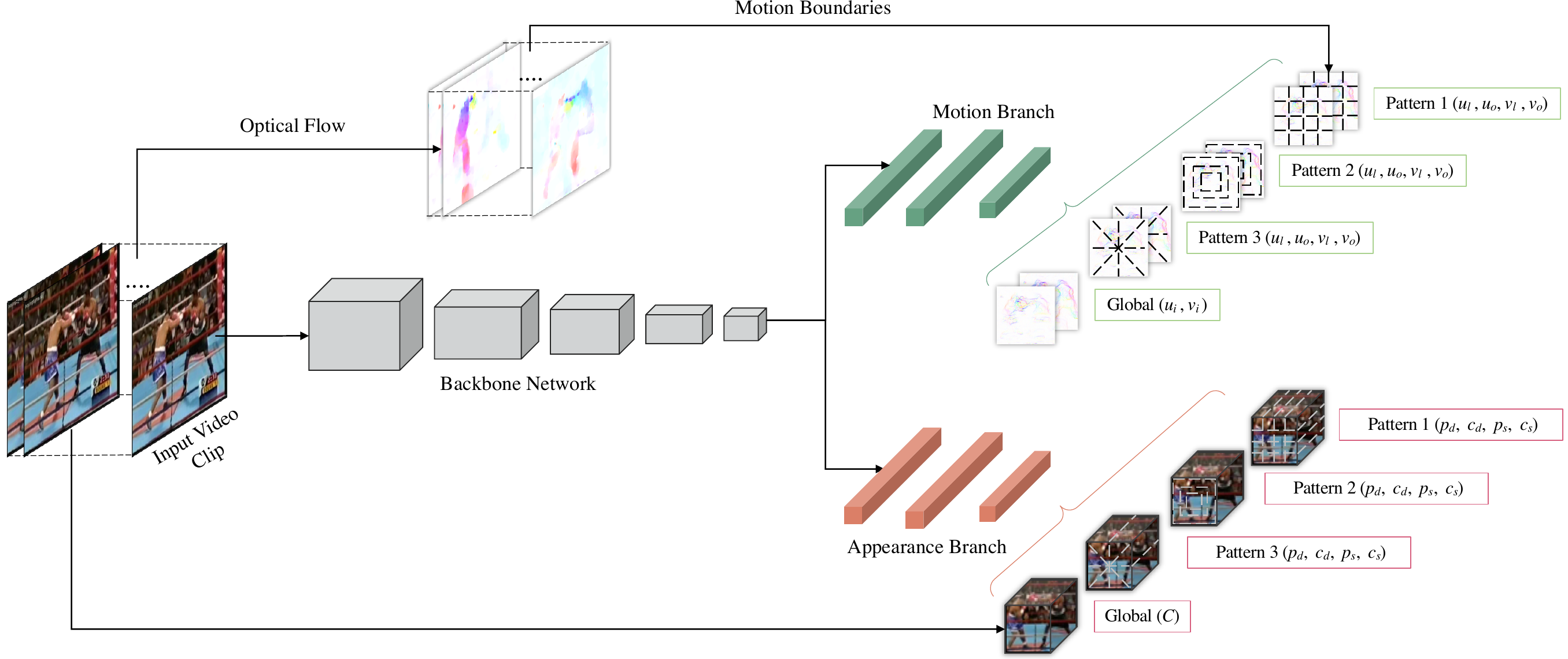}
    \caption{The network architecture of the proposed method. Given a 16-frame video, we regress 14 outputs for the motion branch and 13 outputs for the appearance branch. For each motion pattern, 4 labels are generated by aggregating motion boundaries $M_u$ and $M_v$: (1) $u_l$ -- the largest magnitude location of $M_u$. (2) $u_o$ -- the corresponding orientation of $u_l$. (3) $v_l$ -- the largest magnitude location of $M_v$. (4) $v_o$ -- the corresponding orientation of $v_l$. For each appearance pattern, 4 labels are predicted: (1) $p_d$ -- the position of largest color diversity. (2) $c_d$ -- the corresponding dominant color. (3) $p_s$ -- the position of smallest color diversity. (4) $c_s$ -- the corresponding dominant color. }
    \label{fig:network}
\end{figure*}

\subsection{Appearance Statistics}\label{sec.rgbstat}
\paragraph{Spatio-temporal Color Diversity Labels.} Given an  $N$-frame video clip, same as motion statistics, we divide it into several video blocks by patterns described above.   
For an $N$-frame video block, we first compute the 3D distribution $V_i$ in 3D color space of each frame $i$. 
We then use the Intersection over Union (IoU) along temporal axis to quantify the spatio-temporal color diversity as follows:

\begin{equation}
IoU_{score}=\frac{V_1\cap V_2 \cap ... \cap V_i ... \cap V_N}{V_1\cup V_2 \cup ... \cup V_i ... \cup V_N}.   
\end{equation}
 
The largest color diversity location is the block with the smallest $IoU_{score}$, while the smallest color diversity location is the block with the largest $IoU_{score}$. In practice, we calculate the $IoU_{score}$ on R,G,B channels separately and compute the final $IoU_{score}$ by averaging them.

\vspace{-3mm}
\paragraph{Dominant Color Labels.} After we compute the largest and smallest color diversity locations, the corresponding dominant color is represented by another two labels. In the 3-D RGB color space, we evenly divide it into 8 bins. 
For the two representative video blocks, we assign each pixel a corresponding bin number by its RGB value, and the bin with the largest number of pixels is the dominant color.

\vspace{-3mm}
\paragraph{Global Appearance Statistical Labels.} We also design a global appearance statistics to provide supplementary information. Particularly, we use the dominant color of the whole video as the global statistics. The computation method is the same as described above.  

\subsection{Learning with Spatio-temporal CNNs}\label{sec.learnc3d}
We adopt the popular C3D network \cite{tran2015learning} as the backbone for video spatio-temporal representation learning. Instead of using 2D convolution kernel $k \times k$, C3D proposed to use 3D convolution kernel  $k \times k \times k$ to learn spatial and temporal information together. To have a fair comparison with other self-supervised learning methods, we use the smaller version of C3D as described in \cite{tran2015learning}. It contains 5 convolutional layers, 5 max-pooling layers, 2 fully-connected layers and a soft-max loss layer in the end to predict the action class, which is similar to CaffeNet \cite{jia2014caffe}. We followed the same video pre-processing procedure as C3D. Input video samples are first split into non-overlapped 16-frame video clips.  And for each input video clip, it is first reshaped into 128 $\times$ 171 and then randomly cropped into 112 $\times$ 112 for spatial jittering. Thus, the input size of C3D is 16 $\times$ 112 $\times$ 112 $\times$ 3. Temporal jittering is also adopted by randomly flipping the whole video clip horizontally. 

We model our self-supervised task as a regression problem. The whole framework of our proposed method is shown in Figure \ref{fig:network}. When pre-training the C3D network with the self-supervised labels introduced in the previous section, after the final convolutional layer, we use two branches to regress motion statistical labels and appearance statistical labels separately. For each branch, two fully connected layers are used similarly to the original C3D model design. And we replace the final soft-max loss layer with a fully connected layer, with 14 outputs for the motion branch and 13 outputs for the appearance branch. Mean squared error is used to compute the differences between the target statistics labels and the predicted labels.

\section{Experiments}

In this section, we evaluate the effectiveness of our proposed approach. We first conduct several ablation studies on the local and global, motion and appearance statistics design. Specifically, we use motion statistics as our auxiliary task and appearance statistics acts the similar way. The activation based attention map of different video samples is visualized to validate our proposed methodology.  Second, we compare our method with other self-supervised learning auxiliary tasks on action recognition problem based on two popular dataset UCF101 \cite{soomro2012ucf101} and HMDB51 \cite{kuehne2011hmdb}. Our method achieves the state-of-the-art result. Finally, we conduct two more experiments on action similarity \cite{kliper2012action} and dynamic scene recognition \cite{derpanis2012dynamic} to validate the transferability of our self-supervised spatio-temporal features.

\subsection{Datasets and Evaluations}

In our experiment, we incorporate five datasets: the UCF101~\cite{soomro2012ucf101}, the Kinetics~\cite{kay2017kinetics}, the HMDB51~\cite{kuehne2011hmdb}, the ASLAN~\cite{kliper2012action}, and the YUPENN~\cite{derpanis2012dynamic}. Unless specifically state, we use UCF101 dataset for our model pre-training.

UCF101 dataset \cite{soomro2012ucf101} consists of 13,320 video samples, which fall into 101 action classes. Actions in it are all naturally performed as they are collected from YouTube. Videos in it are quite challenging due to the large variation in human pose and appearance, object scale, light condition, camera view and \etc It contains three train/test splits and in our experiment, we use the first train split to pre-train C3D.

Kinetics-400 dataset is a very large human action dataset \cite{kay2017kinetics} proposed recently. It includes 400 human action classes, with 400 or more video clips for each class. Each sample is collected from YouTube and is trimmed into a 10-seconds video clip. This dataset is very challenging as it contains considerable camera motion/shake, illumination variations, shadows, \etc. We use the training split for pre-training, which contains around 240k videos.

HMDB51 dataset \cite{kuehne2011hmdb} is a smaller dataset which contains 6766 videos and 51 action classes. It also consists of three train/test splits. In our experiment, to have fair comparison with others, we use HMDB51 train split 1 to finetune the pre-trained C3D network and test the action recognition accuracy on HMDB51 test split 1.

 When pre-training on UCF101 train split 1 video data, we set the batch size to 30 and use the SGD optimizer with learning rate 0.001. We divide the leaning rate every 5 epochs by 10. The training process is stopped at 20 epochs. When pre-training on the Kinetics-400 train split, the batch size is 30 and we use the SGD optimizer with learning rate 0.0005. The learning rate is divided by 10 for every 7 epochs and the model is also trained for 20 epochs. When finetuning the C3D, we retain the conv layers weights from the pre-trained network and initialize three fully-connected layers. The entire network is finetuned with SGD on 0.001 learning rate. The learning schedule is the same as the pre-training procedure. When testing, average accuracy for action classification is computed on all videos to obtain the video-level accuracy.

\begin{table}[t]
\caption{Comparison the performance of different patterns of motion statistics for action recognition on UCF101.}
\vspace{-6pt}
\begin{center}
\begin{tabular}{lc}
\hline
Initialization & Accuracy (\%) \\
\hline
Random & 45.4 \\
Motion pattern 1 & 53.8 \\
Motion pattern 2 & 53.2 \\
Moiton pattern 3 & 54.2\\
\hline
\end{tabular}
\end{center}\vspace{-7mm}
\label{pattern}
\end{table}

\subsection{Ablation Analysis}
In this section, we analyze the performance of our local and global statistics, motion and appearance statistics on extensive experiments. Particularly, we first pre-train the C3D using different statistics design on UCF101 train split 1. For local and global statistics ablation studies, we finetune the pre-train model on UCF101 train split 1 data with human annotated labels. For the high-level appearance and motion statistics studies, we also finetune the C3D with HMDB51 train split 1 to get more understanding of the design.

\vspace{-3mm}
\paragraph{Pattern.} The objective of this section is to investigate the performance of different pattern design. Specifically, we use the motion statistics and appearance statistics follow the same trend. As shown in Table \ref{pattern}, all the three patterns outperform the random initialization, \ie, train from scratch setting, by around 8\%,  which strongly proves that our motion statistics is a very useful task. The performance of the three patterns are quite similar, indicating that we have balanced pattern design.

\vspace{-3mm}
\paragraph{Local \emph{v.s.} Global.} In this section, we compare the performance of local statistics, \textit{where is the largest motion video block?},  global statistics, \textit{where is the largest motion frame?} and their combination. As can be seen in Table \ref{global}, only global statistics serves as a useful auxiliary task for action recognition problem, with a improvement of 3\%. And when all the three motion patterns are combined together, we can further get around 1.5\% improvement, compared with single pattern. Finally, all motion statistics labels can achieve 57.8\% accuracy, which is a significant improvement compared with train from scratch. 

\vspace{-3mm}
\paragraph{Motion, RGB, and Joint Statistics.} We finally compare all motion statistics, all RGB statistics, and their combination on UCF101 and HMDB51 dataset as shown in Table \ref{table:joint}.  From the table, we can find that both the appearance and motion statistics serve as a useful self-supervised signals for UCF101 and HMDB51 dataset. The motion statistics is more powerful as the temporal information is more important for video understanding. It is also interesting to note that although UCF101 only improves 1\% when combined motion and appearance, the HMDB51 dataset benefits a lot from the combination, with a 3\% improvement.

\begin{table}[t]
\caption{Comparison of local and global motion statistics for action recognition on the UCF101 dataset.}
\vspace{-6pt}
\begin{center}
\begin{tabular}{lc}
\hline
Initialization & Accuracy (\%) \\
\hline
Random & 45.4 \\
Motion global & 48.3 \\
Motion pattern all & 55.4\\
Motion pattern all + global & 57.8\\
\hline
\end{tabular}
\end{center}\vspace{-4mm}
\label{global}
\end{table}

\begin{table}
\caption{Comparison of different supervision signals on the UCF101 and the HMDB51 datasets.}
\vspace{-6pt}
\begin{center}
\begin{tabular}{lcc}
\hline
Domain & UCF101 acc.(\%) & HMDB51 acc. (\%)\\
\hline
From scratch & 45.4 & 19.7\\
Appearance & 48.6 & 20.3\\
Motion & 57.8 & 29.95\\
Joint & 58.8 & 32.6 \\
\bottomrule
\end{tabular}
\end{center}\vspace{-6mm}
\label{table:joint}
\end{table}

%
\subsection{Action Recognition}
In this section, we compare our method with other self-supervised learning methods on the action recognition problem. Particularly, we compare the results with RGB video input and directly quote the number from \cite{gan2018geometry}. As shown in Table \ref{comparison}, our method can achieve significantly improvement compared with the state-of-the-art both on UCF101 and HMDB51. 
Compared with methods that are pre-trained on UCF101 dataset, we improve 9.3\% accuracy on HMDB51 than \cite{gan2018geometry} and 2.5\% accuracy on UCF101 than \cite{lee2017unsupervised}.
Compared with the method proposed recently \cite{kim2018self} that are pre-trained on Kinetics dataset using 3D CNN models, we can also achieve 0.6\% improvement on UCF101 and 5.1\% improvement on HMDB51. 
And please note that \cite{kim2018self} used various regularization techniques during pre-training, such as channel replication, rotation with classification and spatio-temporal jittering while we do not use these techniques.
The results strongly support that our proposed predicting motion and appearance statistics task can really drive the CNN to learn powerful spatio-temporal features. And our method can generate multi-frame spatio-temporal features transferable to many other video tasks.

\begin{table}[t]
\caption{Comparison with the state-of-the-art self-supervised video representation learning methods on UCF101 and HMDB51.}
\vspace{-18pt}
\begin{center}
\begin{adjustbox}{max width=\columnwidth}
\begin{tabular}{lcc}
\toprule
Method & UCF101 acc.(\%) & HMDB51 acc.(\%) \\
\midrule
DrLim \cite{hadsell2006dimensionality}   & 38.4 & 13.4 \\
TempCoh \cite{mobahi2009deep}   & 45.4 & 15.9 \\
Object Patch \cite{wang2015unsupervised}& 42.7 & 15.6 \\
Seq Ver.\cite{misra2016shuffle} & 50.9 & 19.8 \\
VGAN \cite{vondrick2016generating}& 52.1 & -  \\
OPN \cite{lee2017unsupervised}  & \underline{56.3} & 22.1 \\
Geometry \cite{gan2018geometry} & 55.1 & \underline{23.3}\\
\textbf{Ours (UCF101)}   & \textbf{58.8} & \textbf{32.6} \\
\midrule
ST-puzzle (Kinetics) \cite{kim2018self} & 60.6 & 28.3\\ 
\textbf{Ours (Kinetics)}   & \textbf{61.2} & \textbf{33.4} \\
\bottomrule
\end{tabular}
\end{adjustbox}
\end{center}\vspace{-4mm}
\label{comparison}
\end{table}

\begin{figure}[t]
\begin{center}
   \includegraphics[width=\columnwidth]{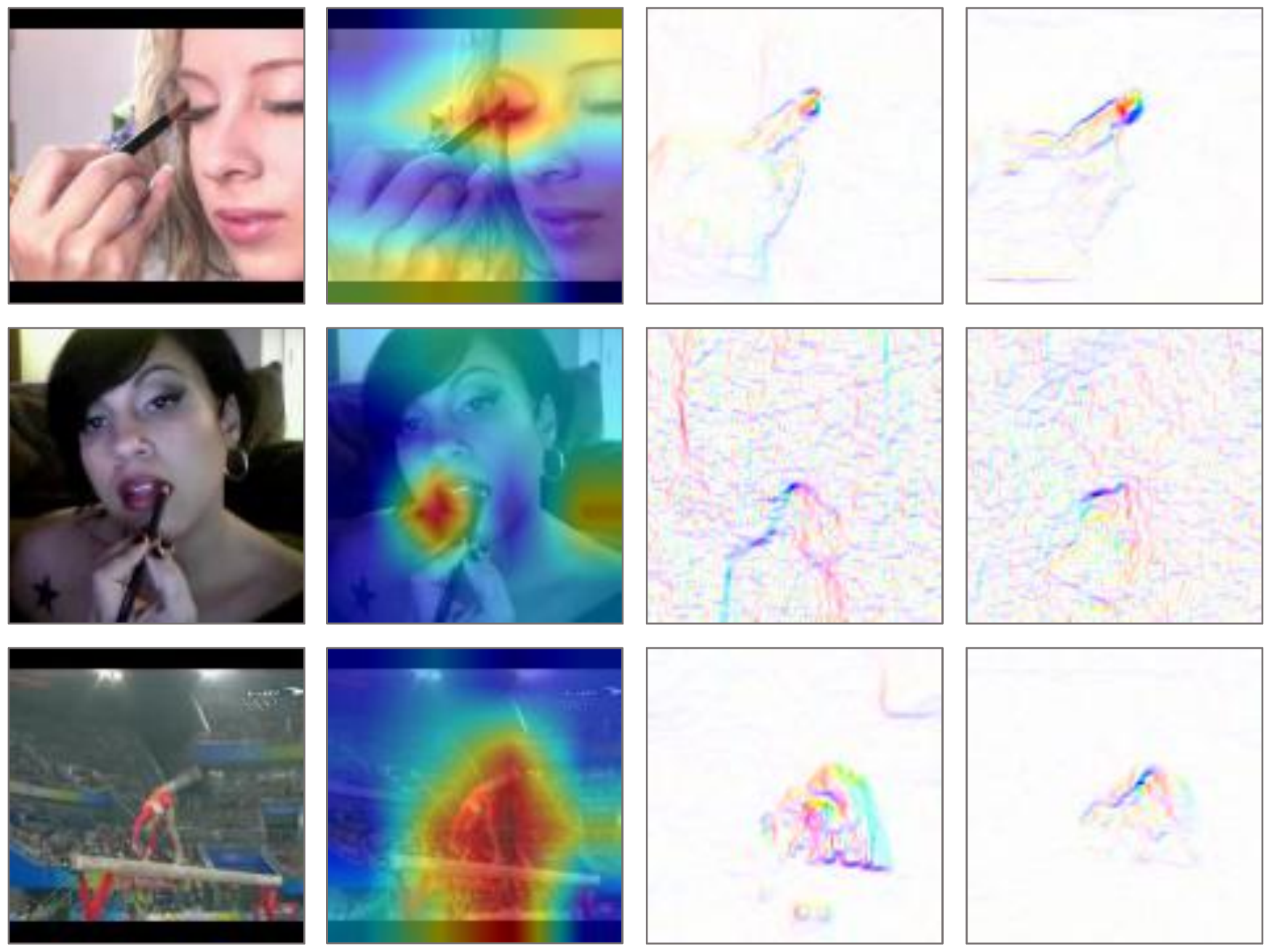} 
\end{center}\vspace{-3mm}
   \caption{Attention visualization. From left to right: A frame from a video clip, activation based attention map of conv5 layer on the frame by using \cite{zagoruyko2016paying}, motion boundaries $M_u$ of the whole video clip, and motion boundaries $M_v$ of the whole video clip.}
\label{fig:attention}\vspace{-4mm}
\end{figure}

\vspace{-3mm}
\paragraph{Visualization.} To further validate that our proposed method really helps the C3D to learn video related features, we visualize the attention map \cite{zagoruyko2016paying} on several video frames as shown in Figure \ref{fig:attention}. It is interesting to note that for similar actions: \textit{Apply eye makeup} and \textit{Apply lipstick}, C3D is just sensitive to the location that is exactly the largest motion location as quantified by the motion boundaries as shown in the right. For different scale motion, for example, the \textit{balance beam} action, the pre-trained C3D is also able to focus on the discriminative location.  

\subsection{Action Similarity Labeling}
We validate our learned spatio-temporal features on ASLAN dataset \cite{kliper2012action}. This dataset contains 3,631 video samples of 432 classes. The task is to predict whether the given two videos are of the same class or not. We use C3D as a feature extractor, followed by a \textit{linear} SVM to do the classification. Each video sample is split into several 16 frames video clips with 8 frames overlapped and then go through a feed-forward pass on C3D to extract features from the last conv layer. The video-level spatio-temporal feature is obtained by averaging the clip feature, followed by l2-normalization. When testing on the ASLAN dataset, we follow the same 10-fold cross validation with leave-one-out evaluation protocol in each fold. Given a pair of videos, we first extract C3D feature from each video and then compute the 12 different distances described in \cite{kliper2012action}. The 12 (dis-)similarity are finally concatenated together to obtain a video-pair descriptor which is then fed into a linear SVM classifier. 
Since the scales of each distance are different, we normalize the distances separately into zero-mean and unit-variance as described in \cite{tran2015learning}. 

As no previous self-supervised learning methods have done experiment on this dataset, to validate that our self-supervised task can drive C3D to learn powerful spatio-temporal features, we design 4 scenarios to extract features from ASLAN dataset: (1) Use the random initialization C3D as feature extractor. (2) Use the C3D pre-trained on UCF101 with labels as feature extractor. (3) Use the C3D pre-trained on UCF101 with our self-supervised task as feature extractor. (4) Use the C3D finetuned on UCF101 on our self-supervised model as feature extractor. Table \ref{ASLAN} shows the performance of different feature extractors. The random initialization model can achieve 51.4\% accuracy as the problem is a binary classification problem. What surprises us is that although our self-supervised pre-trained C3D has never seen the ASLAN dataset before, it can still do well in this problem and outperforms the C3D trained with human-annotated labels by 1.1\%. Such results strongly support that our proposed self-supervised task is able to learn powerful and transferable spatio-temporal features. This can be explained by the internal characteristics of the action similarity labeling problem. Different from the previous action recognition problem, the goal of ASLAN dataset is to predict video similarity instead of predicting the actual label. To achieve good performance, C3D must understand the video context, which is just what we try to drive the C3D to do with our self-supervised method. When finetuned our self-supervised pre-trained model with labels on UCF101, we can further get around 3\% improvement. It outperforms the handcrafted features STIP \cite{kliper2012action}, which is the combination of three popular features: HOG, HOF, and HNF (a composition of HOG and HOF).     

\begin{table}
\caption{Comparison with different handcrafted features and our proposed four scenarios performance on the ASLAN dataset.}
\vspace{-6pt}
\begin{center}
\begin{tabular}{lc}
\hline
Features & Accuracy (\%) \\
\hline
HOF \cite{kliper2012action} & 56.68 \\
HOG \cite{kliper2012action} & 59.78 \\
STIP \cite{kliper2012action} & 60.9 \\
\hline
C3D, random initialization & 51.7 \\
C3D, train from scratch with label & 58.3 \\
\textbf{C3D, self-supervised training} & \textbf{59.4} \\
C3D, finetune on self-supervised & 62.3 \\
\hline
\end{tabular}
\end{center}\vspace{-7mm}
\label{ASLAN}
\end{table}

\subsection{Dynamic Scene Recognition}
The performance on UCF101, HMDB51 and ASLAN dataset shows that our proposed self-supervised learning task can drive the C3D to learn powerful spatio-temporal features for action recognition problem. One may wonder that can action-related features be generalized to other problems? We investigate this question by transferring the learned features to the dynamic scene recognition problem based on the YUPENN dataset \cite{derpanis2012dynamic}, which contains 420 video samples of 14 dynamic scenes. For each video in the dataset, first split it into 16 frames clips with 8 frames overlapped. The spatio-temporal features are then extracted based on our self-supervised C3D pre-trained model from the last conv layer. The video-label representations are obtained by averaging each video-clip features, followed with l2 normalization. A \textit{linear} SVM is finally used to classify each video scene. We follow the same leave-one-out evaluation protocol as described in \cite{derpanis2012dynamic}. 

We compared our methods with both hand-crafted features and other self-supervised learning tasks as shown in Table \ref{YUPENN}. Our self-supervised C3D outperforms both the traditional features and self-supervised learning methods. It shows that although our self-supervised C3D is trained on a action dataset, the learned weights has impressive transferability to other video-related tasks.

\begin{table}
\begin{center}
\caption{Comparison with hand-crafted features and other self-supervised representation learning methods for dynamic scene recognition problem on the YUPENN dataset.}
\vspace{-6pt}
\begin{adjustbox}{max width=\columnwidth}
\begin{tabular}{lcccccc}
\toprule
Method & \cite{feichtenhofer2013spacetime} & \cite{derpanis2012dynamic} & \cite{wang2015unsupervised} & \cite{misra2016shuffle} & \cite{gan2018geometry} & \textbf{Ours} \\
\midrule
Accuracy (\%)  &  86.0 & 80.7 & 70.47 & 76.67 & 86.9 & \textbf{90.2}\\
\bottomrule
\end{tabular}
\label{YUPENN}
\end{adjustbox}
\end{center}\vspace{-7mm}
\end{table}

\section{Conclusions}

In this paper, we presented a novel approach for self-supervised spatio-temporal video representation learning by predicting a set of statistical labels derived from motion and appearance statistics. Our approach is bio-inspired and consistent with human visual systems. We demonstrated that by pre-training on unlabeled videos with our method, the performance of C3D network is improved significantly over random initialization on the action recognition problem. Compared with other self-supervised representation learning approaches, our method achieves state-of-the-art performances on UCF101 and HMDB51 datasets. 
This strongly supports that our method can drive C3D network to capture more crucial spatio-temporal information. We also showed that our pre-trained C3D network can be used as a powerful feature extractor for other tasks, such as action similarity labeling and dynamic scene recognition, where we also achieve state-of-the-art performances on public datasets.

\vspace{-3mm}
\paragraph{Acknowledgements:} This work is supported in part by the Natural Science Foundation of China under Grant U1613218 and 61702194, in part by the Hong Kong ITC under Grant ITS/448/16FP, and in part by the VC Fund 4930745 of the CUHK T Stone Robotics Institute. Jianbo Jiao is supported by the EPSRC Programme Grant Seebibyte EP/M013774/1.

{\small

}

\end{document}